\documentclass[aps,prl,twocolumn,superscriptaddress,showpacs]{revtex4-2}
\usepackage{graphicx}
\usepackage{array}
\usepackage{amsfonts}
\usepackage{amssymb,amsmath,multirow,rotate}
\usepackage{mathrsfs}
\usepackage{booktabs}
\usepackage{threeparttable}
\usepackage{multirow}
\usepackage{epsfig}
\usepackage{threeparttable}
\usepackage{chngpage}
\usepackage{latexsym}
\usepackage{dcolumn}
\usepackage{graphics,graphicx,bm,fleqn,epic,eepic,float}
\usepackage{verbatim}
\usepackage{subfigure}
\usepackage{color}
\usepackage{xr}
\usepackage{float}
\usepackage{listings} 
\usepackage{placeins} 
\usepackage{tikz}
\usepackage{url}
\usepackage{soul}

\definecolor{red}{rgb}{1,0,0}

\definecolor{blue}{rgb}{0,0,1}

\begin{document}

\title{Machine-learning parameter tracking with partial state observation}

\date{\today}

\author{Zheng-Meng Zhai}
\affiliation{School of Electrical, Computer and Energy Engineering, Arizona State University, Tempe, AZ 85287, USA}

\author{Mohammadamin Moradi}
\affiliation{School of Electrical, Computer and Energy Engineering, Arizona State University, Tempe, AZ 85287, USA}

\author{Bryan Glaz}
\affiliation{Army Research Directorate, DEVCOM Army Research Laboratory, 2800 Powder Mill Road, Adelphi, MD 20783-1138, USA}

\author{Mulugeta Haile}
\affiliation{Army Research Directorate, DEVCOM Army Research Laboratory, 6340 Rodman Road, Aberdeen Proving Ground, MD 21005-5069, USA}

\author{Ying-Cheng Lai} \email{Ying-Cheng.Lai@asu.edu}
\affiliation{School of Electrical, Computer and Energy Engineering, Arizona State University, Tempe, AZ 85287, USA}
\affiliation{Department of Physics, Arizona State University, Tempe, Arizona 85287, USA}

\begin{abstract}

Complex and nonlinear dynamical systems often involve parameters that change with time, accurate tracking of which is essential to tasks such as state estimation, prediction, and control. Existing machine-learning methods require full state observation of the underlying system and tacitly assume adiabatic changes in the parameter. Formulating an inverse problem and exploiting reservoir computing, we develop a model-free and fully data-driven framework to accurately track time-varying parameters from partial state observation in real time. In particular, with training data from a subset of the dynamical variables of the system for a small number of known parameter values, the framework is able to accurately predict the parameter variations in time. Low- and high-dimensional, Markovian and non-Markovian nonlinear dynamical systems are used to demonstrate the power of the machine-learning based parameter-tracking framework. Pertinent issues affecting the tracking performance are addressed. 

\end{abstract}

\maketitle

The behavior of a nonlinear dynamical system is controlled by its parameters. In 
a real-world environment, the parameters typically change or drift with time. For
example, when an optical sensor system is deployed to an outdoor environment, 
climatic disturbances such as temperature and humidity fluctuations can cause
the geometrical and material parameters of the system to change with time. In
an ecological system, seasonal fluctuations and human influences on the 
environment can induce changes in the parameters underlying the population 
dynamics such as the carrying capacity and species decay rates. Often, due to the 
complex interactions between the system and the environment, the simplistic 
assumption that the parameters drift linearly with time is not valid. Rather, 
the variations of the parameters with time can be complicated. A generic feature 
of nonlinear dynamical systems is that even a small parameter change can lead to 
characteristically different and even catastrophic behaviors. For example, a 
nonlinear system can typically exhibit a variety of bifurcations including a 
crisis~\cite{GOY:1983} at which a chaotic attractor is destroyed and replaced 
by transient chaos~\cite{LT:book}, leading to system collapse. Being able to 
predict or forecast how some key system parameters change with time into the 
future can lead to control strategies to prevent system collapse.

The problem of tracking parameter variations is an inverse problem, which is 
difficult even if an accurate mathematical model of the system is known. Our 
assumption is that the parameter of interest cannot be directly accessed or 
measured, so tracking its variations will need to be done indirectly using the 
measurements of some accessible dynamical variables of the system. Suppose that 
a key parameter will change with time in the future but, at present the system is 
stationary so that a few distinct values of this parameter can be measured, together 
with the time series of a subset of the dynamical variables. A scenario is that an 
instrument or device is to be deployed in certain missions where the harsh and 
nonstationary environment will cause the key parameter to change with time. Before 
deployment, the device can be tested in a controlled laboratory environment where 
the values of the parameter and the corresponding time series can be obtained. 
Assuming in the real environment the parameter cannot be measured but some time 
series from partial state observation still can be, we ask the question of whether 
it is possible to extract the parameter variations. In this Letter, we demonstrate 
that machine learning can be exploited to provide an affirmative answer.

\begin{figure}[ht!]
\centering
\includegraphics[width=\linewidth]{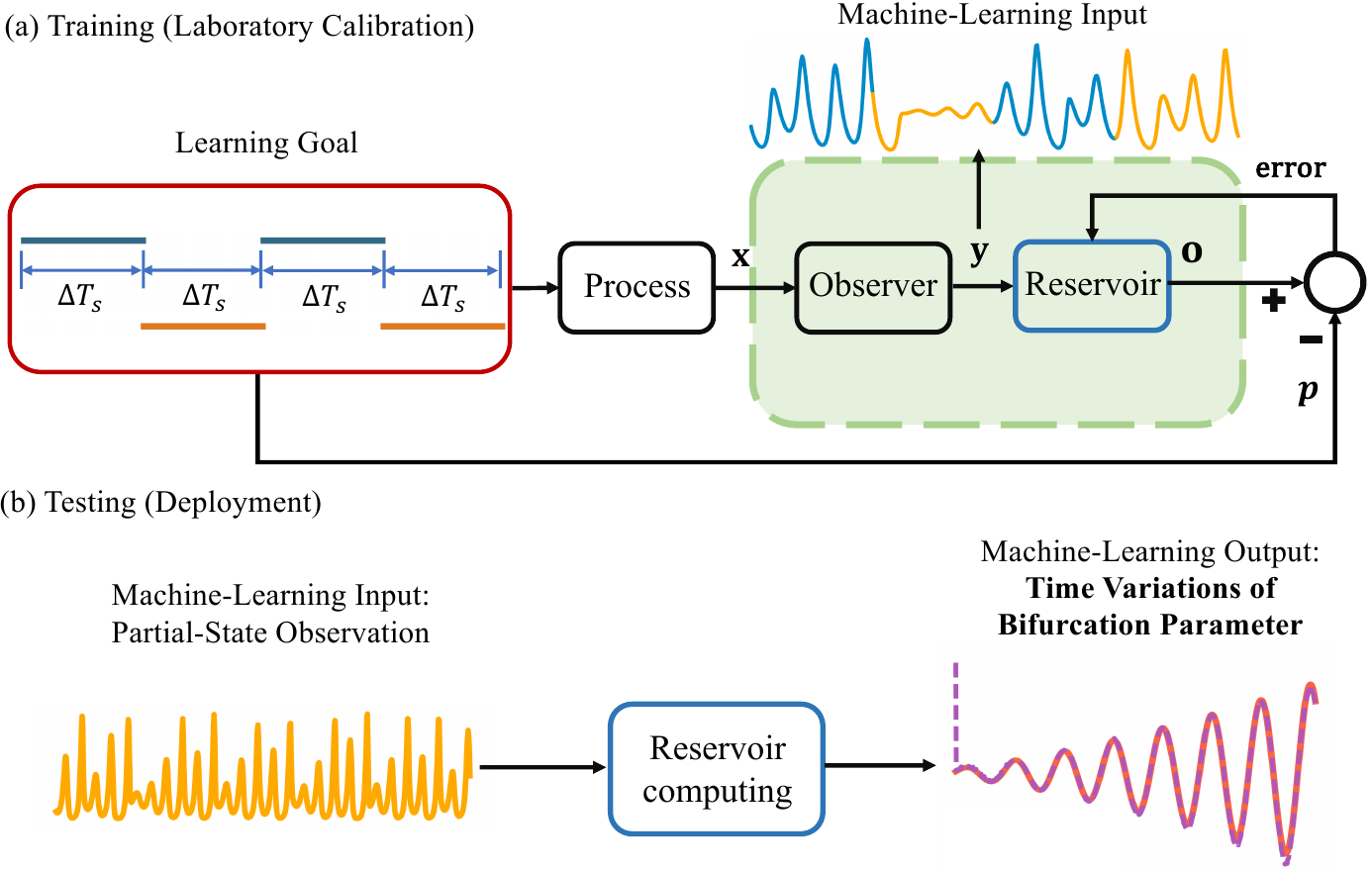} 
\caption{Proposed reservoir-computing based parameter tracking scheme. The goal 
is to track a parameter of the system from which only partial state observation is 
available. The output of the reservoir computer is $o(t)$. (a) In the training 
or laboratory calibration phase, the input to the neural network is an integrated 
vector time series $\mathbf{y}_T(t)$ recombining segments of the time series 
from partial state observation for a small number of parameter values. The 
training goal is the corresponding piecewise constant function of the parameter 
values versus time, denoted as $p(t)$, i.e., minimizing the error between $p(t)$ 
and $o(t)$. (b) In the testing or deployment phase where the parameters are 
inaccessible, the input to the neural network is the current observation vector 
time series $\mathbf{y}(t)$ and the output is the parameter as a function of 
time.}
\label{fig:inverse}
\end{figure}

The idea of exploiting machine learning for parameter tracking has been 
investigated recently~\cite{hannink2016sensor,chen2020machine,chen2020differential,zhang2021neural,abdusalomov2022improved,kao2022deep}
(see Supporting Information~\cite{SI} for an extensive background review
of previous works on parameter identification and tracking). In existing 
machine-learning works, the time series from all dynamical variables and the time 
variations of the parameter is required for training. Here, we articulate a 
machine-learning framework with the following two main features that go beyond 
the existing methods: (1) only the measurements from a partial set of the 
dynamical variables are needed, and (2) observation of the state from a small 
number of parameter values suffices. More specifically, let 
$\mathbf{x} \in \mathcal{R}^D$ be the $D$-dimensional state vector of the 
dynamical system and let $\mathbf{y} \in \mathcal{R}^{D'}$ be the measurement 
or observation vector: $\mathbf{y} = \mathbf{g}(\mathbf{x})$, where $D' < D$ and 
$\mathbf{g}: \mathcal{R}^D \rightarrow \mathcal{R}^{D'}$ is the measurement 
function. We choose reservoir 
computing~\cite{Jaeger:2001,MNM:2002,lukovsevivcius2009reservoir,ASSDMDSMF:2011}
as the machine-learning architecture, which in recent years has been applied to 
predicting nonlinear dynamical systems~\cite{HSRFG:2015,LBMUCJ:2017,PLHGO:2017,LPHGBO:2017,PHGLO:2018,Carroll:2018,NS:2018,ZP:2018,GPG:2019,JL:2019,TYHNKTNNH:2019,FJZWL:2020,ZJQL:2020,KKGGM:2020,KFGL:2021a,PCGPO:2021,KLNPB:2021,FKLW:2021,KFGL:2021b,Bollt:2021,GBGB:2021,Carroll:2022,KWGHL:2023,KB:2023},
and propose the following parameter tracking scheme. Suppose the goal is to track 
a single parameter $p$ (for simplicity), so the output of the neural network is a 
scalar quantity $o(t)$. In a well-controlled laboratory environment, vector time 
series of dimension $D'$ from a small number of parameter values can be measured. 
We construct the input data by first breaking the measured time series into a 
number of segments of equal length and recombining them to form an integrated 
vector time series $\mathbf{y}(t)$. Corresponding to each segment in 
$\mathbf{y}(t)$, there is an exact value of the parameter, thereby generating a 
piecewise constant function of the parameter with time, denoted as $p(t)$. The 
goal of training is to minimize the error between $o(t)$ and $p(t)$, as shown in 
Fig.~\ref{fig:inverse}(a). This arrangement ensures that the neural network learns
the dynamical ``climate'' of the underlying system and how it changes with time 
through alternating exposure to the measurements taking from different parameter 
values. During the testing phase, e.g., when the system is deployed to a real 
application environment, the parameter varies with time and it is no longer 
accessible to observation or measurement. What can be observed is vector time 
series $\mathbf{y}(t)$. When the well-trained neural network takes in 
$\mathbf{y}(t)$ as the input, its output should give the time variation of the 
parameter, realizing accurate parameter tracking, as illustrated in 
Fig.~\ref{fig:inverse}(b).


We consider three prototypical nonlinear dynamical systems: a three-species chaotic
food chain system~\cite{mccann1994nonlinear}, the chaotic R\"{o}ssler 
oscillator~\cite{rossler1976equation}, and the Mackey-Glass delay-differential 
equation system~\cite{mackey1977oscillation}. The first two systems are 
three-dimensional  while the Mackey-Glass system is non-Markovian with an 
infinite-dimensional phase space. For each system, three types of parameter 
variations are considered: frequency modulation (FM), sawtooth wave, and 
amplitude modulation (AM), with different numbers of parameter sampling in 
different ranges for training. The machine-learning performance is characterized 
by the root-mean-square error (RMSE). Here we present the results from the chaotic 
food-chain system given by (results for the other two systems can be found in 
SI~\cite{SI}): $dR/dt = R(1-R/{\rm K}) - {\rm x_c y_c} C R/(R + {\rm R_0})$, 
$dC/dt = {\rm x_c}C [{\rm y_c} R/(R+{\rm R_0})-1] - {\rm x_p y_p} P C/(C+{\rm C_0})$, 
$dP/dt = {\rm x_p}P [{\rm y_p} C/(C + {\rm C_0})-1]$, 
where $R$, $C$, and $P$ are the population densities of the resource, consumer,
and predator species, respectively. The system has seven parameters:
${\rm K,x_c,y_c,x_p,y_p,R_0,C_0}>0$. To illustrate the process of parameter
tracking, we assume that the following three parameters: ${\rm K,y_c,y_p}$ are
time-varying, and fix all other parameters at constant values: $\rm x_c=0.4$,
$x_p=0.08$, $C_0=0.5$, and $R_0=0.16129$, according to some bioenergetics
argument~\cite{mccann1994nonlinear}. We focus on tracking a single parameter:
$\rm{K}$, $\rm{y_c}$, or $\rm{y_p}$, whose respective nominal values are 0.94,
1.7, and 5.0. When tracking one of the three parameters, the other two are fixed
at their respective nominal values.

\begin{figure} [ht!]
\centering
\includegraphics[width=\linewidth]{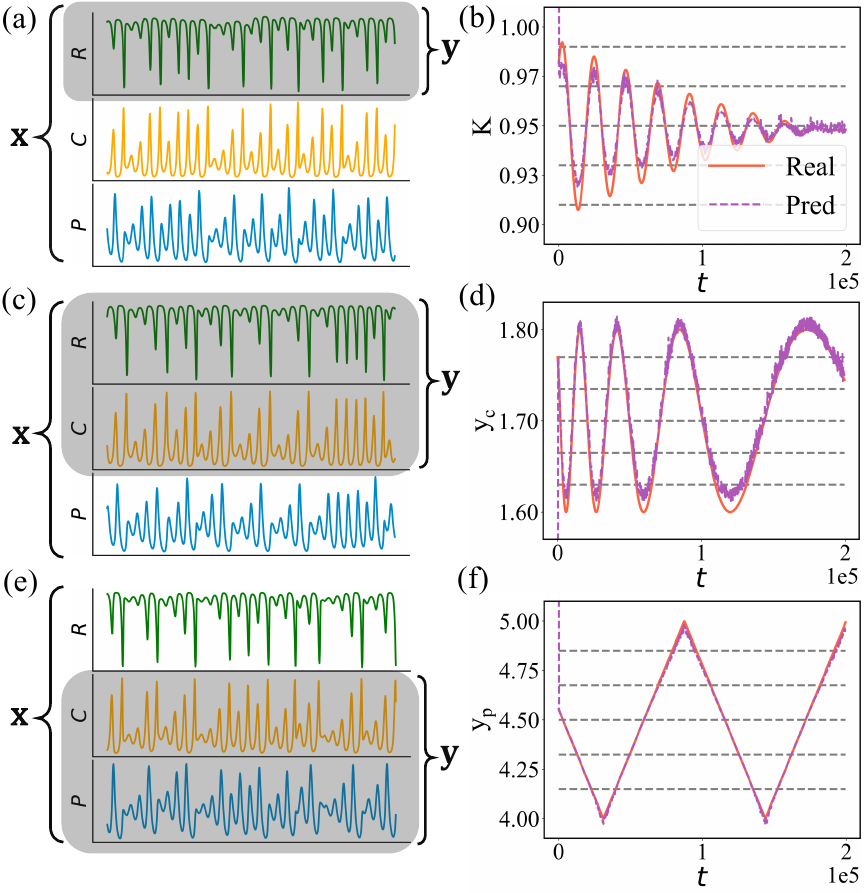}
\caption{Tracking time-varying parameters of the chaotic food-chain system.
Different combinations of the parameter waveforms and partial state observation 
are illustrated: the top, middle, and bottom row correspond to three types of 
parameter variations (AM, FM, and sawtooth waveform), while the gray shaded 
region in the left column illustrates the partial state observation. The right 
column gives the results of parameter tracking in comparison with the ground 
truth. (a,b) One-dimensional observational state variable $\mathbf{y}=(R)$ for 
tracking AM parameter $\rm{K}$; (c,d) Two-dimensional observation 
$\mathbf{y}=(R,C)$ for tracking FM parameter $\rm{y_c}$; (e,f) Two-dimensional 
observation $\mathbf{y}=(C,P)$ for tracking the sawtooth-wave parameter 
$\rm{y_p}$. The five horizontal dashed lines indicate the parameter values from 
which training data are generated, i.e., $s_n=5$.}
\label{fig:partial}
\end{figure}

Figure~\ref{fig:partial} exemplifies the results of tracking the three different
types of parameter variations for different combinations of the bifurcation
parameter and state observation. The computational setting is as follows. Time 
series are generated by integrating the system model using the time step 
$dt=0.01$. The initial states of both the dynamical process and the reservoir 
neural network are randomly chosen from a uniform distribution. The training and 
testing data are obtained by sampling the time series at the interval $\Delta_s$. 
We set $\Delta_s=250dt=2.5$, corresponding to approximately $1/25$ cycles of 
oscillation in the chaotic food-chain system. Let $\Delta T_s$ be the switching 
time interval in which the target parameter is a constant. The training time is 
$900 \Delta T_s$. The testing length is chosen to be slightly longer than the 
training length, enough for tracking several cycles of the parameter variation. 
The size of the reservoir network is $D_r=500$ and the bias number is set to one. 
The other six hyperparameters are determined by Bayesian optimization 
(SI~\cite{SI}), which are fixed during training and testing. Often, a constant 
bias may arise between the machine predicted parameter variation and the ground 
truth, which can be removed by calibrating using two parameter values 
(SI~\cite{SI}). The results in Fig.~\ref{fig:partial} indicate that the 
machine-learning framework is capable of accurate parameter tracking based on 
partial state observation only.

Two characteristics of the bifurcation parameter values used in the training
which can affect significantly the performance of tracking are the number of
such parameter values (denoted as $s_n$) and the relative range of the parameter
variation (denoted as $s_w$) from which the training time series data are
generated. Intuitively, if the training data come from only one value of the
bifurcation parameter, it will not be possible for the reservoir
computer to learn the features of parameter variation, resulting in a large
tracking error. As $s_n$ increases from one, we expect the error to decrease.
What is the minimum number of bifurcation parameter values required for accurate
parameter tracking? Likewise, if the bifurcation parameter values are taken from
the full range of the actual parameter variation, accurate tracking is likely,
resulting in a small error. As the range $s_w$ is reduced, the error will
increase. How tolerant can the machine-learning parameter tracking scheme be 
with respect to this range?

\begin{figure} [ht!]
\centering
\includegraphics[width=\linewidth]{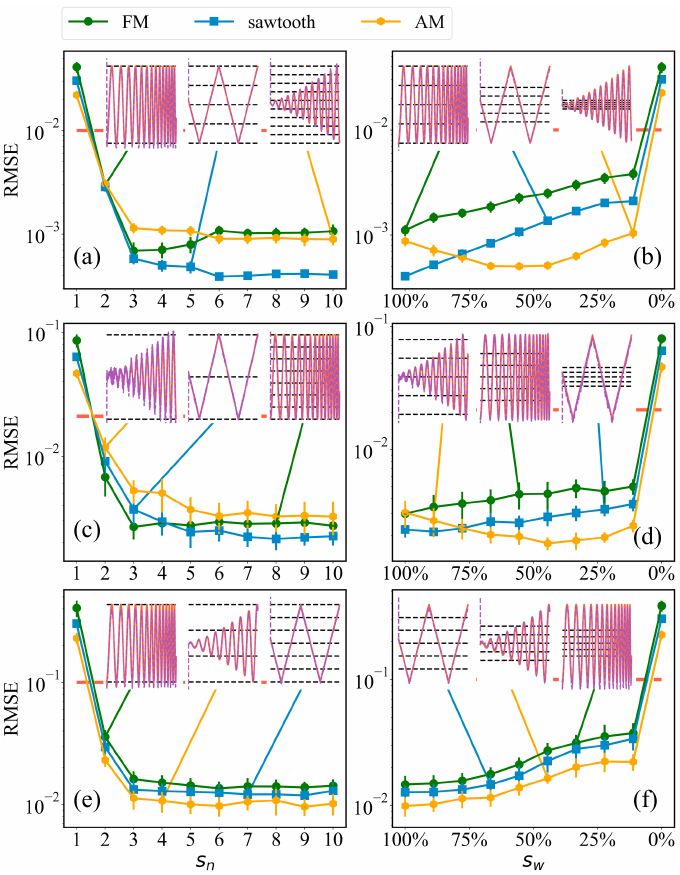}
\caption{Effects of the training bifurcation-parameter values on tracking. The
two relevant quantities are $s_n$ - the number of distinct bifurcation-parameter
values for training, and $s_w$ - the relative parameter range from which time 
series are taken for training. (a,c,e) Testing RMSE versus $s_n$ for parameters
$\rm{K}$, $\rm{y_c}$, and $\rm{y_p}$, respectively. (b,d,f) Testing RMSE versus 
$s_w$ for $\rm{K}$, $\rm{y_c}$, and $\rm{y_p}$, respectively. Each panel has 
results from FM, sawtooth, and AM types of parameter variations, and each point 
is the result of averaging over 50 independent realizations. The results in 
(a,c,e) indicates that using time series from three distinct values of the 
bifurcation parameter suffices to guarantee accurate parameter tracking. The 
results in (b,d,f) suggests acceptable parameter-tracking performance for 
$s_w > 20\%$.}
\label{fig:sample_number}
\end{figure}

Figures~\ref{fig:sample_number}(a,c,e) demonstrate the effect of varying $s_n$ on
the parameter-tracking performance for different combinations of the bifurcation
parameter and its time variations. It can be seen that, for all cases illustrated,
the testing error decreases dramatically as $s_n$ increases from one to three,
and remains approximately constant afterwards, indicating that using the time 
series from as few as three values of the bifurcation parameter suffices for 
accurate tracking of the actual parameter variations. 
Figures~\ref{fig:sample_number}(b,d,f) show the effect of varying the relative 
range $s_w$ on the tracking performance. As $s_w$ decreases from $100\%$, the 
RMSE increases, but it does so in a slow manner until $s_w$ falls below about 
$20\%$, indicating that the machine-learning parameter tracking scheme is 
remarkably tolerant to the range of the bifurcation parameter from which the 
time series for training are generated.

\begin{figure} [ht!]
\centering
\includegraphics[width=0.7\linewidth]{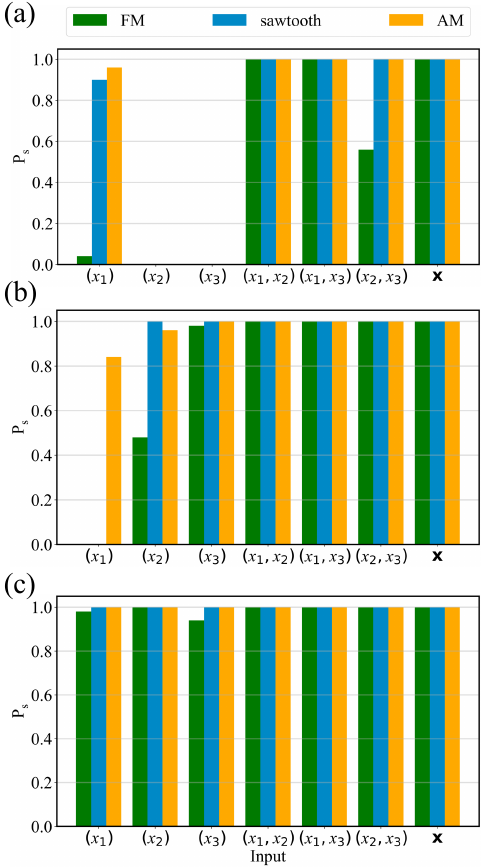}
\caption{Success probability for different partial observation scenarios for the
chaotic food-chain system. (a-c) Bar-graph representation of the success rate for
the seven distinct cases of state observation for the three different bifurcation
parameters, for three types of parameter variation (FM: green; sawtooth: blue;
AM: yellow). The success rate is calculated from 50 independent realizations. In
most cases, observing two state variables can lead to nearly $100\%$ success rate.}
\label{fig:success_rate}
\end{figure}

A key feature of our work is that machine-learning parameter tracking can be done
using only partial state observation. But how partial can the observation be while
still delivering reasonable parameter tracking performance?  For 
$\mathbf{x} \equiv (x_1,x_2,x_3)^T$, full state observation is represented by 
$\mathbf{y} = \mathbf{x}$. There are six cases of partial state observation: 
$\mathbf{y} = (x_1)$, $\mathbf{y} = (x_2)$, $\mathbf{y} = (x_3)$, 
$\mathbf{y} = (x_1,x_2)$, $\mathbf{y} = (x_1,x_3)$, and $\mathbf{y} = (x_2,x_3)$. 
Altogether, there are seven distinct cases of state observation. For each
case, we conduct a reasonable of number (e.g., 50) of tests. For each test, we
define an error threshold. If the resulting testing RMSE is below the threshold,
the test is deemed successful. The fraction of successful cases gives the
success probability. We then calculate, for different types of parameter
variations, the success rate $\rm P_s$ and use a bar graph to represent the rate
for the seven cases of state observation. Some representative results for the
chaotic food-chain system are shown in Fig.~\ref{fig:success_rate}. It can be
seen that in most cases, close to $100\%$ success rate can be achieved by
observing two state variables. Depending on the specific parameter, in some
cases observing even one state variable can lead to satisfactory success rate.

In summary, we developed a reservoir-computing based framework to continuously 
track parameter variations with time based on observation from a subset of the 
full state variables. Relying on partial observation is necessary in real 
applications where not all the dynamical variables of the system are accessible 
and there can be unknown hidden variables. Our unique training scheme enables 
the reservoir computer to learn the correspondence between the input time series 
and the underlying parameter value so that it is able to forecast the parameter 
variations with time based in input time series from partial state observation.
We demonstrated the working of the proposed parameter-tracking framework using 
chaotic systems subject to three distinct types of parameter variations. The 
effects of a number of factors on the tracking performance were investigated: 
minimally required bifurcation parameter values for training, the number of state 
vectors associated with partial-state observation, measurement and dynamical 
noises, network size and training time, and the minimum switching time required 
of the training data (SI~\cite{SI}). The developed reservoir-computing based scheme 
represents a general and robust parameter-tracking framework that can be deployed 
in real-world applications.

A possible application is in epidemiology. For example, some virus has seasonal 
behaviors: its spreading rate varies in different seasons. Our framework can be
used to track the spreading rate change with time based on state evolution, 
thereby enabling accurate prediction of the infection scale and period. Another 
potential application is predicting if a dynamical system is about to approach a 
tipping point at which a transition from a normal to a collapsing (e.g., 
large-scale extinction in an ecosystem) steady state occurs. With inevitable
noises, we can use the noisy time series from partial state observation as the 
input to our machine-learning scheme to generate the time-varying behavior of 
some key bifurcation parameters of the system. Based on the predicted trend of 
the parameter variation, the chance of a tipping point occurring in the near 
future can be assessed. 

This work was supported by the Air Force Office of Scientific Research through Grant No.~FA9550-21-1-0438.


%

\end{document}